\documentclass{article}
\usepackage{spconf,amsmath,graphicx}
\usepackage{color}
\usepackage{url,hyperref,cleveref}
\usepackage{CJKutf8}
\usepackage{enumitem}

\usepackage{makecell}
\usepackage{multicol,multirow}
\usepackage{tablefootnote}
\usepackage{diagbox} 
\usepackage{booktabs}
\usepackage[normalem]{ulem}

\title{Romanization Encoding for Multilingual ASR}
\name{Wen Ding$^*$, Fei Jia$^*$, Hainan Xu, Yu Xi, Junjie Lai, Boris Ginsburg}
\address{NVIDIA Corporation\\
\url{{wend, fjia}@nvidia.com} }
\begin{document}
%
\maketitle
\def\thefootnote{*}\footnotetext{Equal contribution}
\begin{abstract}
We introduce romanization encoding for script-heavy languages to optimize multilingual and code-switching Automatic Speech Recognition (ASR) systems. By adopting romanization encoding alongside a balanced concatenated tokenizer within a FastConformer-RNNT framework equipped with a Roman2Char module, we significantly reduce vocabulary and output dimensions, enabling larger training batches and reduced memory consumption. Our method decouples acoustic modeling and language modeling, enhancing the flexibility and adaptability of the system. In our study, applying this method to Mandarin-English ASR resulted in a remarkable 63.51\% vocabulary reduction and notable performance gains of 13.72\% and 15.03\% on SEAME code-switching benchmarks. Ablation studies on Mandarin-Korean and Mandarin-Japanese highlight our method's strong capability to address the complexities of other script-heavy languages, paving the way for more versatile and effective multilingual ASR systems.
\end{abstract}
\section{Introduction}
\label{sec:intro}
Multilingual Automatic Speech Recognition (ASR) systems are designed to recognize and transcribe speech in multiple languages. 
Code-switching (CS) is a special case of this, dealing with speech that switches between two or more languages within a single utterance or conversation.
While emerging cutting-edge web-scale large speech models such as \cite{radford2023robust, zhang2023google, pratap2023scaling} demonstrate magnificent performance on multilingual ASR, they still fall short in CS scenarios \cite{peng23promptwhisper}, often due to a lack of natural CS data for training. 
This scarcity hinders the ability of both general large speech models and specialized CS ASR systems to effectively learn and integrate acoustic and linguistic information~\cite{haizhou2019lowresource}.

Part of the challenge of multilingual and CS ASR arises from text representations of languages from different language families. 
Languages like those in the Indo-European family usually use a Latin-based alphabet with relatively smaller character sets. 
These can be efficiently represented using methods like byte-pair encoding (BPE), which breaks down words into smaller pieces or sub-words. Research has shown that using sub-words can lead to better performance in language processing tasks \cite{irie2019choice,kudo-2018-subword}.  
However, languages such as Mandarin, Korean, and Japanese have a much larger set of unique characters, making sub-word representation less practical. While there are methods to break these characters into smaller units (like love in Mandarin's 
 \begin{CJK*}{UTF8}{gbsn}爱情\end{CJK*} $\rightarrow$ ài qíng with Pinyin and Korean's \begin{CJK*}{UTF8}{mj}사랑 $\rightarrow$ ㅅㅏ ㄹㅏㅇ\end{CJK*}  with Jamo) and group these characters into sub-units (i.e. ài qàng $\rightarrow$ àiqàng with segmentation), using these phonetic and semantic representations may not always yield the best results \cite{korean_unit,chinese_unit,Meng2019IsWS}.
Despite the effectiveness of character-based approaches for individual languages, their integration into multilingual models is challenging. For instance, \cite{mms70} documents the use of 8k characters for Mandarin, 4k for Japanese, and 2k for Korean, alongside a standardized set of 512 sub-words per language for other languages. This approach yields 11k unique tokens for these three languages alone, leading to a significantly large and potentially imbalanced vocabulary that inflates the model's output dimension.

Effectively encoding languages with unique scripts is crucial for multilingual and CS ASR models.
Many non-Latin languages can be transcribed into the Latin alphabet through romanization.
For instance, pinyin, the primary romanization system for Standard Chinese, facilitates a mapping where a single character can be represented by different pinyins with tones representing the pronunciation.
Typically, around 1,000 distinct pinyins with tones can represent about 5,000 Chinese characters. While romanization doesn't provide a strict one-to-one match, it effectively reduces the vocabulary size and allows the encoder to focus on learning acoustic modeling. 

We propose separating acoustic and language modeling in multilingual and CS ASR models, 
using romanization to reduce vocabulary size and speed up training and inference, aiming to improve model performance and adaptability. 
This approach enhances system flexibility and allows for the use of advanced decoders like Large Language Models (LLMs) for efficient conversion. With this approach, we can utilize synthetic text data for easy fine-tuning to address the shortage of CS audio data. 

In this paper, we make the following contributions:
\begin{itemize}[itemsep=2pt,topsep=0pt,parsep=0pt]
    \item Romanization is investigated to be served as encoding method in multilingual and CS ASR tasks. We apply our encoding method with a balanced concatenated tokenizer to FastConformer-RNNT with a Roman2Char decoder without introducing additional modules such as Language Modeling (LM). 
    \item Experiments on Mandarin-English CS  data show that our model significantly reduces vocabulary size and the dimensions of the output layer, supports larger training batches, lowers memory consumption, and achieves promising outcomes. We release the checkpoints and implementations in NeMo$^1$. 
    \item Our ablation studies on Mandarin-Korean and Mandarin-Japaneses multilingual data demonstrate the effectiveness and generalizability of the proposed method. 
\end{itemize}

\def\thefootnote{1}\footnotetext{\url{https://github.com/wd929/NeMo/tree/code-switch}}

\vspace{-10pt}
\section{Related work}
\vspace{-5pt}
Beyond the method described in \cite{mms70}, OpenAI's Whisper~\cite{radford2023robust} system uses Byte-level Byte Pair Encoding (BBPE) \cite{radford2019language} for text tokenization, proving effective across various applications.
However, it faces challenges with languages that have unique scripts or significantly differ from the Indo-European family, like Hebrew, Chinese, and Korean, primarily due to BBPE's limitations in handling distinct scripts or linguistic structures. 
Research noted in \cite{apple_bbpe} indicates that BBPE can lead to higher deletion rates in bilingual End-to-End (E2E) ASR systems due to invalid byte sequences, and these BBPE-based bilingual systems underperform compared to their monolingual counterparts.  
Google USM~\cite{zhang2023google}'s approach with word-piece models (WPMs) also struggles with script diversity, resulting in large output layers and difficulties in scaling. 
Conversely, for complex-script languages like Chinese, substituting characters with Pinyin for text encoding in Natural Language Processing (NLP) and ASR tasks typically offers greater efficiency and robustness compared to processing each character individually as demonstrated in
 \cite{tacl_subchar_2023, yang2022effectiveness, decoupling, chuang2021non}. 

Romanization Encoding has been studied in both NLP and speech processing fields. The uroman tool, introduced by \cite{hermjakob-etal-2018-box}, converts texts to Latin-scripts, aiming for phonetic representation.
This work has been applied in multilingual pretrained language models \cite{purkayastha2023romanization} to enhance the low-resourced languages.
Uroman is also utilized for pretraining the speech processing system in \cite{pratap2023scaling} as additional forced alignment to tokenize texts. 
Uroman's unidirectional nature poses a challenge for ASR tasks that require original script output and an additional deromanization step. 

Various languages have multiple romanization methods.
While uroman is universal, our focus is on the most popular Romanization systems for each language studied: Pinyin for Chinese, Revised Romanization for Korean, and Hepburn Romanization for Japanese.
This approach aims to preserve maximum phonetic and linguistic information and avoid unnecessary transformations, such as uroman converting digital numbers in various scripts to Western Arabic numerals. 
In addition, as phonological distinctions might be lost during the romanization process making deromanization more difficult \cite{riyadh-kondrak-2019-joint}. Thus it is more feasible to unify the romanization and deromanization procedure in an end-to-end fashion. 


Researchers have explored specialized model architectures \cite{zhou20b_interspeech,lu2020bi} for code-switching tasks to better capture language-specific information.
This includes integrating Language Identification (LID) \cite{shan2019cslid,liu2023reducing,dalmia2021transformer} enhancement strategies and leveraging pre-trained models alongside LM based Beam Search \cite{chuang2021non} during inference to boost performance.
Despite these advancements, the evaluation of these models on monolingual test sets often goes unexamined, and the volume of training data available is typically constrained. This situation is largely attributed to the scarcity of CS data.
To enhance the data and domain scope for CS ASR training, methods such as transfer learning from monolingual ASR to initialize encoders with both monolingual and code-switched datasets have been implemented in \cite{shan2019cslid}. 
Additionally, efforts including synthetic text data generation have been explored to further augment the training resources \cite{chuang2020trainingcs,winata2019code,Li2020ImprovingCL}.
To our knowledge, only one publicly accessible checkpoint$^2$ for Mandarin-English CS exists and it was solely trained on CS dataset SEAME~\cite{lyu2010seame}.  
\def\thefootnote{2}\footnotetext{\url{https://huggingface.co/espnet/vectominist\_seame\_asr\_conformer\_bpe5626}}

\begin{figure}[h]
  \centering
  \includegraphics[width=25em]{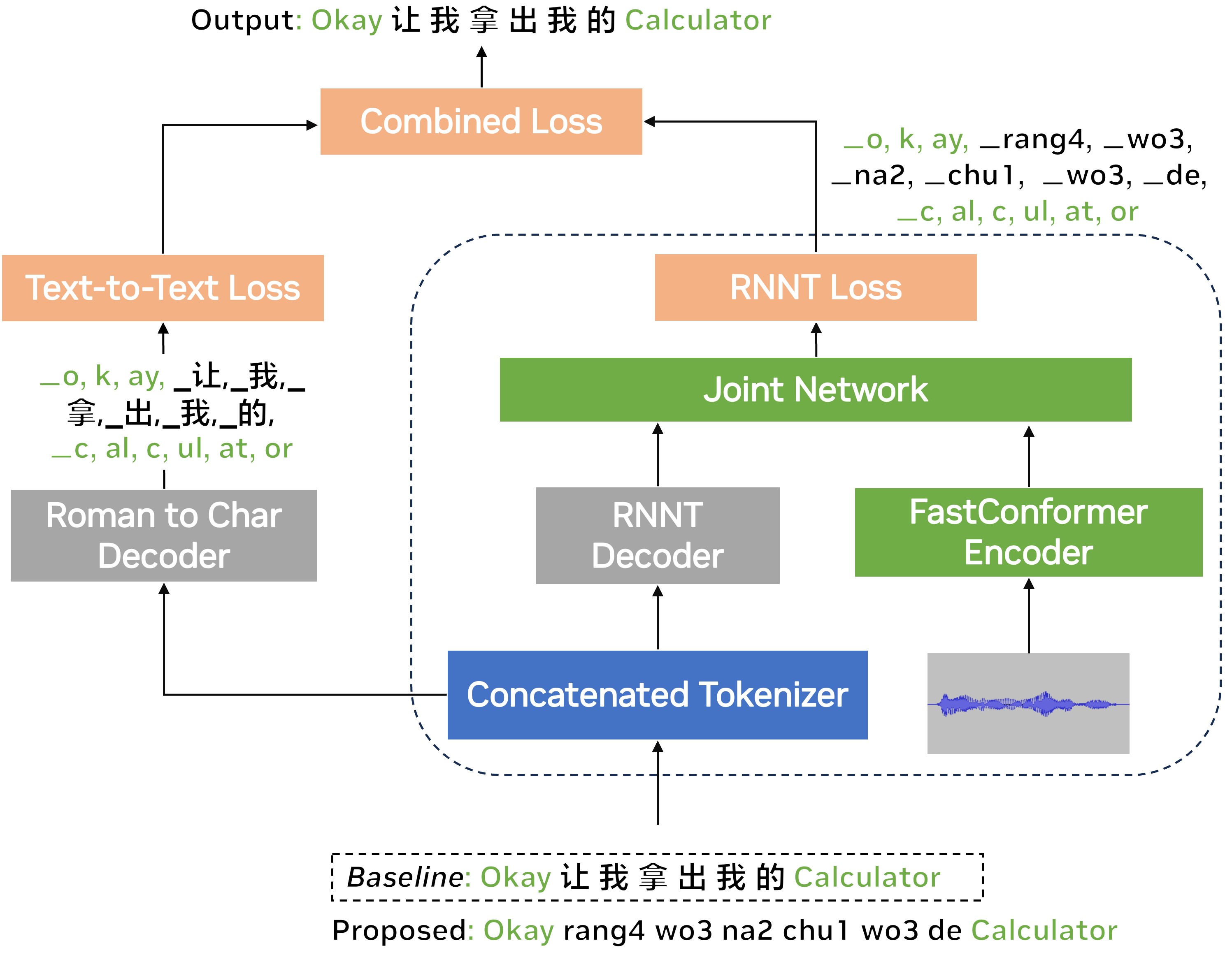}
  \caption{The proposed approach builds upon the baseline Fast-Conformer RNNT model, which incorporates a Concatenated Tokenizer and is outlined within a dashed rectangle.  Instead of using direct Char input/output for Mandarin and BPE for English, our approach applies romanization encoding, feeding Pinyin (for Mandarin) and BPE (for English) into the Fast-Conformer RNNT. The Roman to Char Decoder then maps these inputs back to Char and BPE, respectively. The model is trained end-to-end (E2E), combining text-to-text loss with RNNT loss for optimization.}
  \label{figure:plot_temp}
  \vspace{-1.5em}
\end{figure}
\vspace{-10pt}
\section{Method}
\vspace{-5pt}
\subsection{Model structure}
Our model structure is illustrated in Figure \ref{figure:plot_temp}, where we take a ZH-EN CS as an example.
In this work, we employ the Fast Conformer model  \cite{rekesh2023fast} combined with Recurrent Neural Network Transducer (RNNT) technology \cite{graves2012sequence} as our baseline. 
The Fast Conformer is an optimized version of the original Conformer model. 
It features a new downsampling schema that significantly reduces computational requirements by approximately 2.9 times and enhances inference speed, while maintaining or even improving performance across various Speech and NLP tasks. 
RNNT excels in capturing sequence knowledge and is popular in both monolingual \cite{hainan2023tdt} and CS ASR \cite{dalmia2021transformer} tasks. 
However, despite RNNT's strengths, it tends to struggle with script-heavy languages that have large vocabularies, as it can severely limit batch sizes and slow down training 
\cite{Kuang2022PrunedRF}.

The baseline, highlighted by a dashed rectangle inputs ZH characters and EN words. In contrast, our method replaces characters with Romanized text (Pinyin) before processing through a concatenated tokenizer for text representations (detailed in Section~\ref{subsec:tokenizer}). Meanwhile the audio representation is learned by FastConformer encoders. We introduce another decoder that transcribes the romanized text to characters (described in Section~\ref{subsec:r2c}). 

\subsection{Roman and BPE concatenated tokenizer}
\label{subsec:tokenizer}
Following the concatenated tokenizer approach introduced by \cite{kunal}, we leverage pre-trained monolingual tokenizers to construct a combined tokenizer.
This setup ensures separate label spaces for each language, enabling good generalization capabilities.
For instance, English tokens are assigned indices ranging from [0,1023], while Mandarin tokens are allocated indices starting from 1024 up to 1024 + vocab size.
The aggregated approach, while achieving similar performance levels to the non-aggregated method, offers the added advantage of facilitating LID. It does so by providing pseudo language information during training, thereby enabling the model to learn LID representations internally.

We prepare 1,024 English BPE sub-word units using the LibriSpeech (LS) dataset \cite{dan2015librispeech} and roughly 5,000 Chinese characters from the AISHELL-2 (AS2) dataset \cite{du2018aishell}. 
These Chinese characters were then romanized into Latin characters referred to here as \textit{`Roman'} encoding for all languages using the PyPinyin$^3$ toolkit. 
As seen in Table~\ref{table:stat_tokenizer}, this romanization process reduced the Chinese vocabulary size from 5,178 to 1,239, enabling us to create a balanced concatenated tokenizer for Mandarin and English.
For Korean and Japanese, similar processing methods were employed using the  kroman$^4$ and pykakasi$^5$ toolkit, respectively, to achieve comparable reductions in vocabulary size and to facilitate a unified approach to tokenizer construction across multiple languages.

\def\thefootnote{3}\footnotetext{\url{https://github.com/mozillazg/python-pinyin}}
\def\thefootnote{4}\footnotetext{\url{https://github.com/victorteokw/kroman}}
\def\thefootnote{5}\footnotetext{\url{https://github.com/miurahr/pykakasi}}

\begin{CJK*}{UTF8}{mj}
\begin{table*}[!ht]
\centering
\caption{Examples of romanization for Mandarin, Korean, Japanese and Mandarin-English. For Latin-based languages such as English, we bypass romanization and directly employ Byte Pair Encoding (BPE), setting the vocabulary count at 1,024. }
\vspace{5pt}
\begin{tabular}{lllr|lr}
\hline
&language & Char & vocab  & Roman & vocab\\
\hline
(a)&Mandarin & 差 不 多  & 5178  & cha4 bu4 duo1 &  1239 \\
(b)&Korean  & 안 녕 하 세 요 & 1202 &  an nyeong ha se yo &   1202   \\
(c)&Japanese & か な 漢 字 & 3329 & ka na kan ji & 1059 \\
\hline
(d)&Mandarin-English & 差 不 多 ten minutes  & 6202 & cha4 bu4 duo1 ten minutes  &  2263 \\
\hline
\end{tabular}
\label{table:stat_tokenizer}
\end{table*}
\end{CJK*}

\subsection{Roman to Character Decoder}
\label{subsec:r2c}
To translate the romanized units Roman back to original characters, we introduce an additional module in our system called the Roman to Character (R2C) decoder. 
This Transformer-based model is designed to learn the multi-to-multi mappings between romanized text and original characters. For English, a language that already uses the Latin alphabet, the inputs and outputs remain unchanged as shown in (d) of in Table~\ref{table:stat_tokenizer}.

Importantly, despite the E2E training approach, the R2C decoder functions independently of the RNNT encoder outputs, focusing solely on learning the sequence-to-sequence mapping. 
Only its loss is merged with the RNNT loss, allowing for the possibility of separate training with text data or integration with more advanced pre-trained translation models, such as LLMs. 
This flexibility also enables the module's extension to other languages with complex scripts, like Korean and Japanese.
During training, accurately labeled Roman sequences serve as inputs for the module. 
To streamline the process, the decoding stage exclusively uses the greedy search hypothesis from the RNNT decoder as input, simplifying the overall pipeline.
\section{Experiments}

\subsection{Data}

\textbf{Code-switching data} 
SEAME is a publicly available dataset designed for Mandarin-English speech recognition, containing Mandarin, English, and natural intra-sentential code-switching data from interviews and conversations. The dataset specifics, including the duration of each data type, are outlined in Table~\ref{table:seame_data}. 
``ZH" refers to the monolingual Mandarin segments, and ``EN" to the monolingual English parts. The dataset includes approximately 60 hours of natural CS data, a quantity considered limited for training robust models. Notably, most of SEAME speakers are from Singapore and Malaysia, presenting accents different from Mainland China. For the purposes of model selection, 10\% of the training set samples are randomly chosen to form a validation set.

\textbf{Monolingual data} 
AISHELL-2 is an extensive 1000-hour open-source Mandarin speech corpus, the speakers of which mainly are from Mainland China. 
LibriSpeech comprises 960 hours of English speech from native speakers. These two monolingual datasets are used in our experiments to enhance the performance of code-switching ASR systems.

\textbf{Evaluation data}
For evaluation, we stick to the data division of SEAME established by \cite{zeng2019end}, which includes a test set for Mandarin speech named test\_man and another tailored to Southeast Asian accented English, labeled test\_sge. The specific durations of these test sets are also detailed in Table~\ref{table:seame_data}. 
Additionally, to assess performance on monolingual data, test sets from AISHELL-2 (as2\_test) and LibriSpeech (ls\_clean) are utilized in our analysis.


\begin{table}
\caption{Duration composition of Mandarin (ZH), English (EN), and code-switching (CS) utterances in SEAME corpus. The duration of Mandarin dev sets (as2\_test) and English (ls\_clean) are also included. }
\vspace{5pt}
\begin{resizebox}{1.0\columnwidth}{!}{
\begin{tabular}{lrrrrrr}
\hline
& train & val & test\_man & test\_sge & as2\_test & ls\_clean\\
\hline
duration(h) & 85.4     & 9.8  & 7.5  & 3.9 & 4.0  & 5.4    \\
ZH (\%) & 16.6    &16.3    & 13.3  & 5.1 & 100 & 0        \\
EN (\%)  & 15.8  &16.3    & 6.6  & 41.0   & 0 & 100     \\
CS (\%)  & 67.4    & 67.3   & 80.0   & 53.8 & 0 & 0  \\
\hline
\end{tabular}
}
\end{resizebox}
\vspace{-5pt}
\label{table:seame_data}
\end{table}

\vspace{-10pt}
\subsection{Experiment Setup}
\label{sec:experiment_setup}
To evaluate monolingual test sets, Character Error Rate (CER) is applied to Mandarin, while Word Error Rate (WER) is used for English. For CS test sets, we employ Mixed Error Rate (MER), which incorporates word-level measurements for English and character-level assessments for Mandarin.

In all of our experiments, we use the Adam \cite{adam} optimizer combined with a Cosine Annealing learning rate scheduler, including a warm-up phase of 10,000 steps. 
The learning rate is set to peak at 1.5e-3 and then decrease to a minimum of 1e-6. In addition, we incorporate SpecAug~\cite{park2019specaugment} during training process to enhance model robustness and performance.
Model averaging is employed, and during evaluation, greedy search is utilized without the assistance of any external LM or re-scoring techniques.
The detailed training recipe will be open-sourced in NeMo.

\subsection{Results}
Performance of training solely on SEAME dataset is detailed in Table \ref{table:seame_only_perf}, showing that romanization encoding yields improved results for both the test\_man and test\_sge sets. 
In Table \ref{table:add_as_ls}, we integrate monolingual datasets utilizing their rich acoustic and linguistic content to boost multilingual and CS ASR.
The proposed Roman-based method outperforms the Char-based model in both of the test sets of SEAME, with 10.77\% and 9.43\% MER reductions respectively. 
Further analysis by dividing the CS test sets into Mandarin and English segments underscores the advantages of romanization encoding for both languages.

Moreover, to thoroughly assess the model's proficiency in handling both CS and monolingual scenarios, results for monolingual test sets are presented, indicating an improvement in performance on the monolingual Mandarin test set, albeit with a slight decline on the monolingual English test set.
To balance monolingual and code-switching (CS) data within a fixed 2085-hour training data, we upscale CS data to 285 hours and reduce both AISHELL-2 (AS2) and LibriSpeech (LS) data to 900 hours each. 
This adjustment result in significant MER reductions of 13.72\% and 15.03\% in CS test sets, demonstrating improvements over baseline models. 
Performance variations in monolingual sets were observed, largely due to the differing accents and speaking styles of speakers such as speakers from Singapore and Malaysia versus those from Mainland China \cite{lyu2010seame}.
Nevertheless, the model effectively retains its capability to process monolingual information, often delivering equal or superior performance. 


\begin{table}
\centering
\caption{MERs (\%) on the SEAME dataset reveal that the proposed romanization approach surpasses the character-based baseline by reducing vocabulary size, leading to a more balanced tokenizer and improved performance.}
\vspace{5pt}
\begin{resizebox}{0.8\columnwidth}{!}{
\begin{tabular}{lrrr}
\hline
Encoding & vocab & test\_man & test\_sge \\
\hline
Char+BPE    &   6202 & 22.26    &32.30             \\
Roman+BPE    &  2263 & 21.99     &31.45          \\
\hline
\end{tabular}}
\end{resizebox}
\label{table:seame_only_perf}
\vspace{-10pt}
\end{table}

\begin{table*}[t]
  \centering

   \caption{Adding monolingual AISHELL2 (AS2) and LibriSpeech (LS) data during training. Results are evaluated with CS testsets including test\_man and test\_sge, and monolingual testsets including the testset of AS (as2\_test) and test\_clean of LS (ls\_clean). 
To balance the different acoustic and language information, we attempt upsampling CS data but keep the total number of training data fixed. }
   \vspace{5pt}
  \begin{resizebox}{2.0\columnwidth}{!}
  {
  
    \begin{tabular}{c c |  c  c  c |  c c c  c c c | c | c}
      \toprule
       
           \multicolumn{2}{c|}{Encoding} &\multicolumn{3}{c|}{Dataset (hours)}& \multicolumn{6}{c|}{SEAME} & \multicolumn{1}{c|}{AS2} & \multicolumn{1}{c}{LS}  \\
      \cmidrule(lr){1-2}\cmidrule(lr){3-5}\cmidrule(lr){6-11}\cmidrule(lr){12-12}\cmidrule(lr){13-13}

       \multirow{2}{*}{ZH} & \multirow{2}{*}{EN} & \multirow{2}{*}{SEAME} & \multirow{2}{*}{AS2} & \multirow{2}{*}{LS} & \multicolumn{3}{c}{test\_man} & \multicolumn{3}{c}{test\_sge} &  \multicolumn{1}{|c|}{as2\_test}  & \multicolumn{1}{c}{ls\_clean}   \\
       \cmidrule(lr){6-8}\cmidrule(lr){9-11}\cmidrule(lr){12-12}\cmidrule(lr){13-13}

    ~&~ &~&~ & ~& MER & CER & \multicolumn{1}{c}{WER} & MER & CER & WER & \multicolumn{1}{c|}{CER} & \multicolumn{1}{c}{WER}   \\
    \midrule
    Char & BPE & 85                        & 1000                         & 1000        &      17.64 & 16.87 & 28.08                             &        25.35 & 26.33 & 29.44                           &                  7.74                       &                   \textbf{2.60}                  \\

Roman & BPE&85                       & 1000                          & 1000         &   15.74 & 15.10 & 25.28                               &    22.96 & 21.97 & 27.41                             &       \textbf{7.05}                                  &             3.26 \\ 

Roman & BPE& 285                       & 900                          & 900         & \textbf{15.22} & \textbf{14.79} & \textbf{24.31}                                  & \textbf{21.54} & \textbf{20.74} & \textbf{25.70}                                 &       7.48                                  &          2.75          \\

\bottomrule
\end{tabular}}
\end{resizebox}
\label{table:add_as_ls}
\vspace{-5pt}
\end{table*}

\vspace{-5pt}
\subsection{Ablation Study}
\label{ablation_study}
The proposed romanization encoding approach is designed to be easily adaptable to various languages. 
This section demonstrates the effectiveness of our method with Korean and Japanese, which face challenges such as limited publicly available corpora, insufficient for training large ASR models, and lack of CS data. 
Through training a multilingual model on constraint datasets (less than 50 hours), we demonstrate our approach's capability and efficiency in data-limited situations and its potential extension to other languages.

\textbf{Mandarin-Korean}
\def\thefootnote{6}\footnotetext{\url{https://github.com/goodatlas/zeroth}}
A Mandarin-Korean bilingual ASR model was trained using the entire 50-hour Zeroth Korean$^6$ dataset and 50 hours data randomly selected from the AISHELL-1 dataset \cite{bu2017aishell}. 
Evaluations on monolingual Mandarin (test\_as1) and Korean (test\_zeroth) test sets utilized CERs for performance measurement. 
Results in Table~\ref{table:seame_kr_zh} indicate that the Roman-based bilingual ASR model maintains performance on the Korean test set while achieving better results on the AISHELL-1 test set compared to a character-based model. 

\begin{table}
\centering
\caption{CER (\%) for a Mandarin-Korean Bilingual system trained on 50h+50h of data shows the proposed method reduces Mandarin vocabulary size, boosts its performance, and maintains Korean results.}
\vspace{5pt}
\begin{resizebox}{0.8\columnwidth}{!}{
\begin{tabular}{lrrr}
\hline
Encoding &vocab & test\_as1 & test\_zeroth \\
\hline
Char       &  6380 & 12.87             &1.40  \\
Roman       & 2441 & 12.60          &1.40  \\
\hline
\end{tabular}
}
\end{resizebox}
\label{table:seame_kr_zh}
\vspace{-10pt}
\end{table}

\textbf{Mandarin-Japanese}
We also experiment on Mandarin-Japanese Bilingual ASR, 
drawing training data randomly from 50 hours of the AISHELL-1 dataset and 50 hours from the Japanese ReazonSpeech~\cite{reazonspeech} dataset. 
By using romanization encoding, the concatenated vocabulary size is reduced from 8,507 to 2,298. 
As we can see in Table ~\ref{table:zh-jp-results}, when evaluated on AISHELL1 (test\_as1) and ReazonSpeech (test\_reazon) test sets,  Roman-based model can perform better than Character encoding one in a large margin, which further indicates effeteness and scalability of our proposed Romanization encoding for other script-heavy languages. 

\textbf{Evaluations of R2C module}
Our proposed method includes additional R2C module to transcribe Roman to characters. The total training parameters are at par with the baseline system since the size of RNNT outputs is decreased. 
Although the end-to-end inference speed for the proposed system can not be faster than the character encoding models but the training batch sizes can be set larger, which is essential for the Multilingual ASR model training. For instance, the reduction in the RNNT concatenated vocabulary size in the Mandarin-Korean Bilingual ASR model from 6,380 to 2,441, primarily due to Mandarin (as Korean mapping is nearly one-to-one), allowed for at least a 2X larger training batch size and more than 20\% quicker RNNT inference compared to models using character encoding. We believe that this work could benefit not only the RNNT-based model but also the popular auto-regressive speech large foundational models. 

\begin{table}
\centering
\caption{CERs (\%) for Mandarin-Japanese Bilingual ASR models indicates Roman encoding reduces vocabulary size by 73\% and significantly enhances performance for both Mandarin and Japanese.}
\vspace{5pt}
\begin{resizebox}{0.8\columnwidth}{!}{
\begin{tabular}{lrrr}
\hline
Encoding &vocab & test\_as1 & test\_reazon \\
\hline
Char       &  8507 & 19.75        &36.00  \\
Roman       & 2298 & 11.30        &29.31  \\
\hline
\end{tabular}
}
\end{resizebox}
\label{table:zh-jp-results}
\end{table}
\section{Conclusion}
In this study, we introduce romanization encoding as a strategy to enhance multilingual ASR systems, particularly for languages with complex scripts. 
Our experiments with Mandarin-English CS ASR illustrate that employing a balanced tokenizer by romanized characters can lead to significant performance gains, with improvements of 13.71\% and 15.03\% on SEAME CS test sets. 
Additionally,  we have extended the application of Roman-based tokenizers to Mandarin-Korean and Mandarin-Japanese multilingual ASR systems, yielding promising results in terms of both faster training speeds and improved performance.
Looking ahead, we plan to refine our approach by applying BPE or similar encoding methods to romanized text, aiming for a more compact and efficient vocabulary. 
Enhancing the R2C decoder with advanced models like LLMs could significantly boost overall accuracy. 
The system's flexibility enables the use of synthetic text data to improve R2C decoder meanwhile leveraging the pre-trained audio encoder to mitigate the limited availability of code-switching audio data.

\newpage

\bibliographystyle{IEEEbib}
\bibliography{citations/refs}

\end{document}